# ROMANIAN DIACRITICS RESTORATION USING RECURRENT NEURAL NETWORKS

STEFAN RUSETI[1], TEODOR-MIHAI COTET[1], AND MIHAI DASCALU[1,2]

[1] *University Politechnica of Bucharest, Computer Science Department*
[2]*Academy of Romanian Scientists*
stefan.ruseti@cs.pub.ro, teodor_mihai.cotet@stud.acs.upb.ro, mihai.dascalu@cs.pub.ro

## *1. Introduction*

Writing texts with Romanian diacritics is in general a cumbersome endeavor as most computer keyboards do not have special keys for specific Romanian letters with diacritics. Hence, people sometimes write digital text in Romanian language with the simplified version of letters with diacritics, namely the ASCII version, thus removing all diacritics. Diacritics change the morphology of a word ('cană' and 'cana' meaning 'cup' and 'the cup', respectively) or even the complete meaning of the word ('fată' and 'față' meaning 'girl' and 'face', respectively); thus, this is a particularly important task for various NLP processes. Therefore, diacritics restoration is a mandatory step for adequately processing Romanian texts, and not a trivial one, as you generally need context in order to proper restore a character.

## *2. Related work*

Most previous methods which were experimented for Romanian restoration of diacritics do not use neural networks. Among those that do, there are no solutions specifically optimized for this particular language (i.e., they were generally designed to work on many different languages). Generally, n-gram based models were tried (Ungurean, Burileanu, Popescu, Negrescu, & Dervis 2008, Petrică, Cucu, Buzo, & Burileanu 2014).

## *3. Method*

### *3.1 Corpus*

The corpus (PAR) contains transcriptions of the parliamentary debates in the Romanian Parliament[1], from 1996 to 2017. The corpus is very diverse in terms of subjects, as it contains the debates of the parliamentary committees in which economic, social, political and judicial issues are discussed exhaustively. The corpus contained around 50M words.

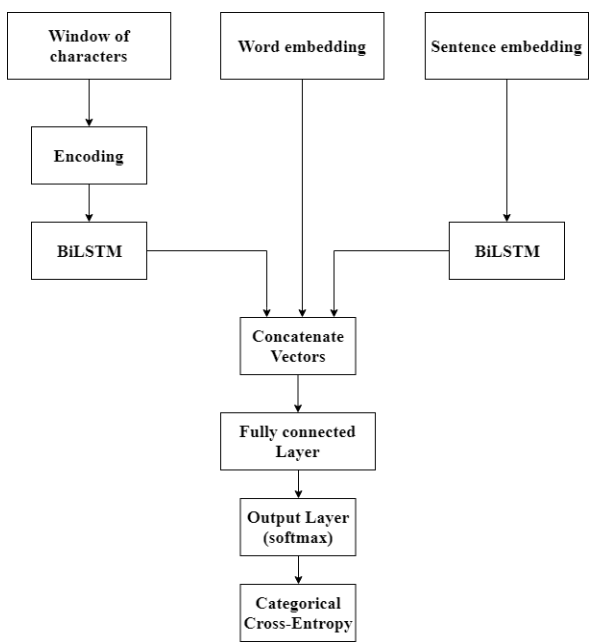

### *3.2 Network Architecture*

The architecture of the proposed model is presented in Figure 1. The neural network is composed on three different paths: characters, current word, and the current sentence. The main idea behind this architecture is to combine lexical with semantic information, therefore capturing more complex contexts.

The first path in the network is represented by a Bidirectional Long Short-Term Memory (BiLSTM; Graves & Schmidhuber, 2005) encoder using character embeddings. A fixed size window is used to represent the context of the current character. The network can learn different similarities between letters by using character embeddings. Also, applying the encoder at the character level allows the network to generalize different forms of the same word, which will have very similar representations, and also generalize for unseen words, if they look similar to other known concepts.

However, the model can benefit from semantic information, as well. In order to achieve this, another path was added to the architecture, represented by a BiLSTM encoder applied on the current sentence. The words in the sentence are represented by pre-trained FastText embeddings. The assumption was that the two encoders are complementary one to another, and the model can learn how to combine the information from both.

The third path is represented by the embedding of the current word. This was added in order to help the network select which part of the encoded context is useful for the current entry. Experiments were also done with models without this word path, which proved to be less accurate.

### *3.3 Input Processing*

For the word embeddings, we have used the pre-trained versions from FastText[2] which were trained by using the skip-gram model (Bojanowski, Grave, Joulin, & Mikolov, 2016). Moreover, we have averaged across all words (i.e. words that are in FastText vocabulary) that match the initial word (matching means that stripping the diacritics you get the initial words) because we did not know exactly the word in case without diacritics and thus, neither its corresponding embedding. Consequently, for each letter of the diacritics-free text (i.e. "a", "e", "i", "s", "t") which is fed into the network, we create these 3 parts.

---

[1] https://www.senat.ro/, http://www.cdep.ro/
[2] https://github.com/facebookresearch/fastText/blob/master/pretrained-vectors.md

## 4. Results

Some of the hyperparameters of the models were not included in Table 2 because they were not modified during the experiments. The cell size of the word BiLSTM encoder was set to 300, the same as the word embedding size. The character window size was set to 13, while the maximum sentence length was 31. The training was done in batches of 256, using the cross-entropy loss function and the Adam optimizer (Kingma & Ba, 2015).

**Table 2**: Char/word accuracy

| Model | Char Embedding | Char LSTM | Hidden | Epochs | Dev char acc (%) | Test char acc (%) | Test word acc (%) |
|---|---|---|---|---|---|---|---|
| Chars | 16 | 32 | 32 | 5 | 98.865 | 98.864 | 97.413 |
| Chars | 20 | 64 | 256 | 5 | 99.012 | 99.017 | 97.750 |
| Chars (5 classes) | 16 | 32 | 32 | 5 | 99.048 | 99.068 | 97.867 |
| Chars | 24 | 64 | 64 | 4 | 99.064 | 99.057 | 97.856 |
| Chars + sentence | 20 | 64 | 256 | 3 | 99.068 | 99.065 | 97.881 |
| Chars + word | 20 | 64 | 256 | 4 | 99.309 | 99.329 | 98.453 |
| Chars + word + sentence | 20 | 64 | 256 | 5 | 99.365 | **99.378** | **98.573** |
| Chars + word + sentence | 20 | 64 | 256, 128 | 5 | **99.380** | 99.366 | 98.553 |

A detailed analysis of the best models, that includes precision and recall values for each letter, was also performed. In Table 3 we present these results for the All-256-128 model that uses characters, words, and sentences with two hidden layers of sizes 256 and 128.

**Table 3**: Detailed performance per letter

| Model | Letter | Precision (%) | Recall (%) | F-Score (%) |
|---|---|---|---|---|
| All-256-128 | "a" | 99.16 | 98.86 | 99.01 |
| | "ă" | 96.29 | 97.31 | 96.80 |
| | "â" | 99.17 | 98.80 | 98.99 |
| | "i" | 99.97 | 99.96 | 99.97 |
| | "î" | 99.65 | 99.72 | 99.69 |
| | "s" | 99.84 | 99.84 | 99.84 |
| | "ș" | 99.44 | 99.43 | 99.43 |
| | "t" | 99.84 | 99.77 | 99.80 |
| | "ț" | 98.97 | 99.29 | 99.13 |

## 5. Discussion

The results presented before show that adding contextual information from the sentence encoded with a BiLSTM can improve the results of a char-based approach significantly. However, only adding the encoded sentence, without the current word is not enough, the accuracy of the char model, and char+sentence being very similar.

The detailed analysis shows that the model is biased towards choosing the class with no diacritics for each letter. Virtually all measures (precision, recall) are higher for letters with no diacritics ("a", "i", "s", "t") compared to the corresponding ones with diacritics. This can be explained by the higher number of letters with no diacritics compared to the ones with diacritics, but also by the missing diacritics in the corpus.

Most of the hardest words to be restored are those in which the diacritics make the word (noun or adjective) indefinite as in "politică", "importanța" and "prezența", which means the model still fails to distinguish between definite and indefinite words.

## 6. Conclusions

In this paper, we proposed a novel neural network architecture that combines lexical information at the character level, with a semantic representation of the surrounding context computed at the word level, using recurrent neural networks. Our experiments show a significant improvement of the accuracy when adding contextual information.

The results show that contextual information helped, but the improvement obtained compared to a neural network that uses the character encoder and the current word is not extremely large. One future improvement could consist in adding an attention mechanism based on the current word, used to better select what is relevant from the context for the current letter. Another observed issue is the imbalance of diacritics in texts (e.g., "a" is much more frequent than "ă" or "â"), which could be solved by using a weighted loss that takes into account these distributions.